%% file: ReviewTemplate.tex
\documentclass[10pt,twocolumn,letterpaper]{article}

\usepackage[pagenumbers]{cvpr} % To force page numbers, e.g. for an arXiv version

\usepackage{graphicx}
\usepackage{amssymb}
\usepackage{booktabs}

\usepackage{amsmath,amsfonts,amsthm,bm}
\usepackage{mathtools}

\usepackage{caption}
\DeclareCaptionLabelFormat{andtable}{#1~#2 \& \tablename~\thetable}
\usepackage[accsupp]{axessibility}
\usepackage[pagebackref,breaklinks,colorlinks]{hyperref}

\usepackage[capitalize]{cleveref}
\crefname{section}{Sec.}{Secs.}
\Crefname{section}{Section}{Sections}
\Crefname{table}{Table}{Tables}
\crefname{table}{Tab.}{Tabs.}

\usepackage[
  font = small,
  tableposition = top
]{caption}

\usepackage{setspace}
\def\cvprPaperID{3195} % *** Enter the CVPR Paper ID here
\def\confName{CVPR}
\def\confYear{2022}
\begin{document}

\title{
% RNNPose: Recurrent 
% 6-DoF Object Pose Refinement  
% via Correspondence Field Estimation and Matching 
% RNNPose: Recurrent 6-DoF Object Pose Refinement 
% Based on Correspondence Field Estimation
% RNNPose: A Robust 6-DoF Object Pose Refiner with Recurrent Correspondence Field Estimation and Pose Optimization
% RNNPose: A Robust Recurrent 6-DoF Object Pose Refiner with  Correspondence-field-based Pose Optimization
RNNPose: Recurrent 6-DoF Object Pose Refinement with Robust Correspondence Field Estimation and Pose Optimization
}

\author{
Yan Xu$^1$~~Kwan-Yee Lin$^1$~~Guofeng Zhang$^2$~~Xiaogang Wang$^1$~~Hongsheng Li$^1$ \\ 
$^1$The Chinese University of Hong Kong~~~~$^2$State Key Lab of CAD\&CG, Zhejiang University\\
{\tt\small yanxu@link.cuhk.edu.hk, zhangguofeng@cad.zju.edu.cn, \{jylin,xgwang,hsli\}@ee.cuhk.edu.hk}
}
\maketitle

% \begin{spacing}{0.945}
%%%%%%%%% ABSTRACT
\begin{abstract}
  % Directly estimating the 
  6-DoF object pose estimation from a monocular image is  challenging, and a post-refinement procedure is generally needed for high-precision estimation. 
  In this paper, we propose a framework based on a recurrent neural network (RNN) for object pose refinement, which is robust to erroneous initial poses and occlusions. During the recurrent iterations, object pose refinement is formulated as a non-linear least squares problem based on the estimated correspondence field (between a rendered image and the observed image). The problem is then solved by a differentiable Levenberg-Marquardt (LM) algorithm enabling end-to-end training. 
  The correspondence field estimation and pose refinement are conducted alternatively in each iteration to recover the object poses. Furthermore, to improve the robustness to occlusion, we introduce a consistency-check mechanism based on the learned descriptors of the 3D model and observed 2D images, which downweights the unreliable correspondences during pose optimization. Extensive experiments on LINEMOD, Occlusion-LINEMOD, and YCB-Video datasets validate the effectiveness of our method and demonstrate state-of-the-art performance. 
  \let\thefootnote\relax\footnotetext{\vspace{-2ex} K. Lin and H. Li are the co-corresponding authors. }
\end{abstract}
  
%%%%%%%%% BODY TEXT
\section{Introduction}
\label{sec:intro}
\input{doc/intro.tex}

% \begin{spacing}{0.91}
\section{Related Work}
\label{sec:related work}
\input{doc/related_work}
% \end{spacing}

\section{Method}
\label{sec:method}
\input{doc/method}
\section{Experiments}
\label{sec:experiment}
\input{doc/experiment}

% \end{spacing}

%-------------------------------------------------------------------------

% Update the cvpr.cls to do the following automatically.
% For this citation style, keep multiple citations in numerical (not
% chronological) order, so prefer \cite{Alpher03,Alpher02,Authors14} to

%------------------------------------------------------------------------

%-------------------------------------------------------------------------

%-------------------------------------------------------------------------

%-------------------------------------------------------------------------

%%%%%%%%% REFERENCES
{\small
\bibliographystyle{ieee_fullname}
\bibliography{egbib}
}

\end{document}

%% file: doc/intro.tex
\input{fig/highlight.tex}
6-DoF object pose estimation is of crucial importance in various applications, including augmented reality, robotic manipulation, and autonomous driving. 
Influenced by varying illuminations and occlusions, appearances of the differently posed objects may vary significantly from different views, which poses great challenges for 6-DOF object pose estimation from a single color image. 

The recent top-performing methods \cite{XiangSNF18,sundermeyer2018implicit,li2018deepim,manhardt2018deep,zakharov2019dpod} additionally include a pose refinement procedure which substantially improves the performance. 
Some of these frameworks \cite{XiangSNF18,sundermeyer2018implicit} rely on depth sensors and refine the poses with the ICP algorithm~\cite{arun1987least}.  
% based on the geometric structures captured by depth sensors. 
To avoid the expensive depth sensor,  Li \etal \cite{li2018deepim} and Manhardt \etal \cite{manhardt2018deep} pioneered the RGB-based pose refinement. 
% by harnessing 
% the regression capability of 
% convolutional neural networks (CNNs).  
% the feasibility and competitiveness of RGB-based pose refinement by exploiting the regression power of CNN. 
% In each refinement iteration, 
During refinement, these methods first render a reference color image according to the coarse pose estimate. 
This rendered image along with the observed image is then fed to a CNN to directly predict the residual pose for refining the coarse pose \cite{li2018deepim,manhardt2018deep,zakharov2019dpod}. 
% between the coarse pose and the observed pose  
% with a CNN, 
While these methods perform well in ideal scenarios based on massive training data, the pose regression becomes less stable in practice. 
More recently, Iwase \etal \cite{iwase2021repose} formulated the object pose refinement as an optimization problem  
based on feature alignment, 
% by aligning the deep features, 
and reported significant performance improvements.  
In their work, the encoded features of a 3D model by a neural network are projected to the 2D image plane according to the pose parameters. 
% as a reference. 
Thereafter, the pose optimization is conducted by aligning the projected features with the observed target image features.
% the deep features of a 3D model are projected to 2D image plane as the reference image, and the pose residual is estimated by aligning the projected features with the observed target image features.  
As the pose optimization 
depends on the gradients from the pixel-level feature differences, the feature alignment based methods are only applicable to small inter-frame pose variations \cite{Younes2019TheAO} and are not quite robust with erroneous initial poses. 
% As the feature alignment process depends on the gradients from the pixel-level feature differences for pose optimization, it is only applicable to small inter-frame pose variations \cite{Younes2019TheAO} and is not quite robust with erroneous initial poses.  when the initial poses are far away from the ground-truth. 
% directly compares the features from the reference image and target image
% reference-image features with the target-image features in pixel-level and refine the pose with the gradients from 
% feature differences,  it is only applicable to small inter-frame pose variations \cite{Younes2019TheAO} and is thus not quite robust when the initial poses are far away from the ground-truth. 
Moreover, Iwase \etal \cite{iwase2021repose} still have a limited design for occlusion handling, which might limit the deployment scope.

In this work, we propose a recurrent object pose refinement framework, dubbed RNNPose, which is robust to erroneous initial poses and occlusions. 
The overall pipeline is illustrated in Fig.~\ref{fig:highlight}. Before refinement, a reference image of the object is rendered according to the initial pose estimation. 
Our refinement module refines the initial pose based on this rendered image and the observed image. 
To increase the tolerance to erroneous initial poses, our refinement is conducted within a recurrent framework, where the pose optimization is formulated as a non-linear least squares problem based on estimated correspondence fields.  
In each recurrent iteration, the dense correspondences between the rendered image and observed image are estimated, and the object pose is then optimized to be consistent with the  correspondence field estimation.  
The architecture of our correspondence estimation is inspired by the recent optical flow estimation techniques \cite{sun2018pwc,teed2020raft}, which is integrated with our pose optimization recurrently. 
To suit our task where unpatterned objects and  illumination variations are ubiquitous, we further include a correspondence field rectification step in each recurrent iteration based on the currently optimized pose. The inconsistent correspondences are rectified by enforcing rigid-transformation constraints. 
The rectified correspondence field is also used to initialize the next recurrent iteration to improve the robustness further. 
% The next recurrent iteration is initialized with the rectified correspondence field to improve the robustness further. 

% To handle the occlusions, 
For occlusion handling, we introduce a 3D-2D hybrid network trained with a contrastive loss, 
% based on \cite{sun2020circle}, 
which generates distinctive point-wise descriptors for the 3D object model and observed 2D images. 
A similarity score is constructed for each estimated correspondence pair based on the learned descriptors, with which to downweight the unreliable correspondences during pose optimization.
The pose optimization is conducted by a differentiable Levenberg-Marquardt (LM) algorithm (sharing the ideas of \cite{tang2018ba,teed2019deepv2d}) for end-to-end training.
% , which enhances the feature learning for correspondence estimation. 

Our contributions are three-fold: 1) We propose an RNN-based 6-DoF pose refinement framework that is robust to large initial pose errors and occlusions.  
During recurrent iterations, the pose optimization is formulated as a non-linear least squares problem based on the estimated correspondence field. Meanwhile, the correspondence field is also being rectified and improved by the optimized pose for robustness.   
2) To handle the occlusions, a 3D-2D hybrid network is introduced to learn point-wise descriptors which are used to downweight unreliable correspondence estimations during pose optimization.   
3) We achieve new state-of-the-art performances on LINEMOD, Occlusion LINEMOD, and YCB-Video datasets. 
Our code is public at \url{https://github.com/DecaYale/RNNPose}.  
% Our code is public at \url{https://github.com/AnonymousRelease/RNNPose}.  

%% file: fig/highlight.tex
\begin{figure}[t!]
    \centering
    \includegraphics[width=0.95\linewidth, trim=2cm 11cm 16cm 0cm, clip]{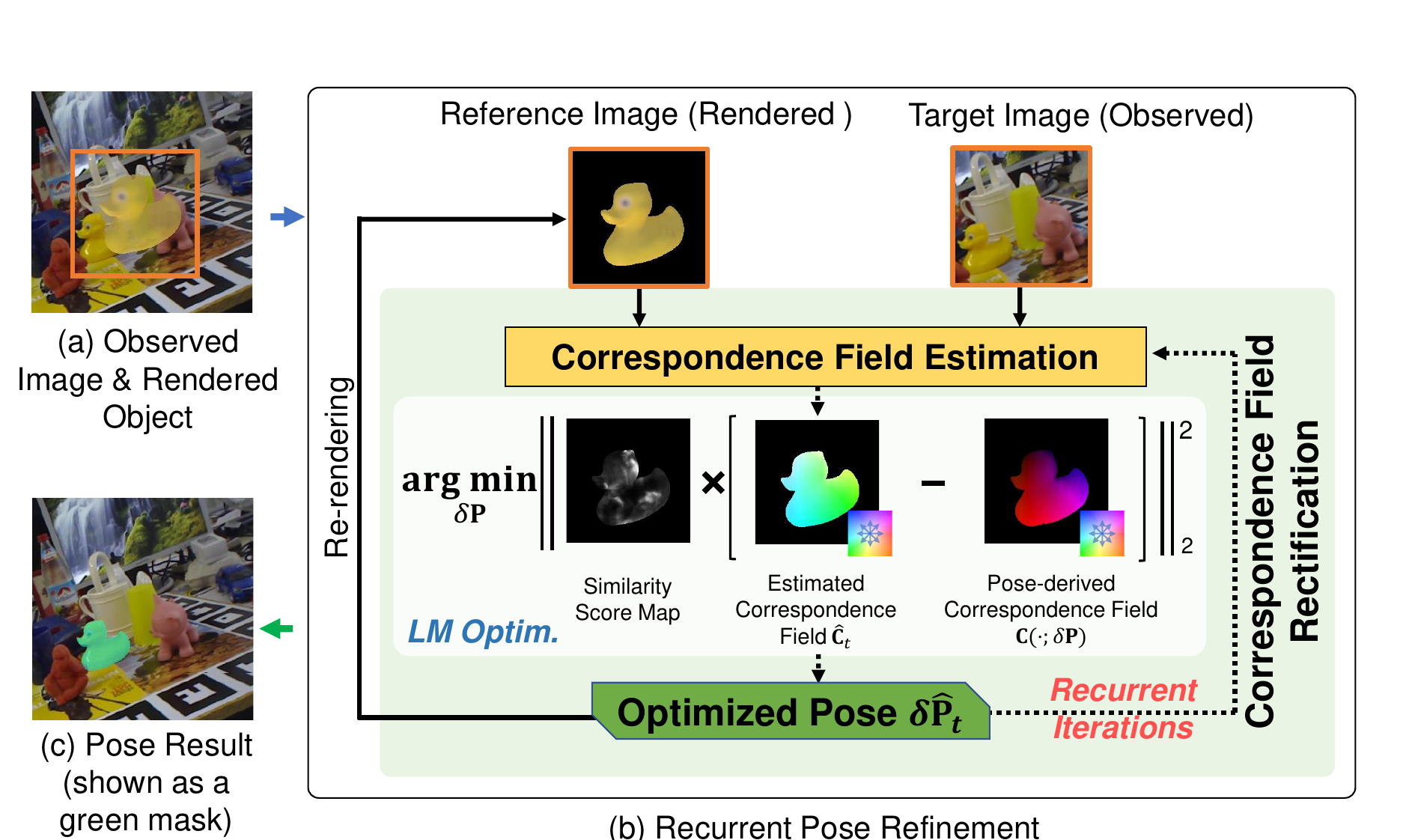}
    \vspace{-0ex}
    \caption{ 
        The basic idea. (a)~Before refinement, a reference image is rendered 
        according to the object initial pose
        % the object CAD model and its initial pose 
        (shown in a fused view).
        (b) Our RNN-based framework 
        % is designed to 
        recurrently refines the object pose based on the estimated correspondence field between the reference and target images. The pose is optimized 
        % by solving a non-linear least squares problem 
        to be consistent with the reliable correspondence estimations highlighted by the similarity score map (built from learned 3D-2D descriptors) via differentiable LM optimization.  
        % is used to downweight the  unreliable correspondences. 
        % The 6-DOF pose refinement is formulated as an non-linear optimization based on the correspondence field between the reference and target images. 
        % The correspondence field estimation and pose optimization are conducted alternately in the RNN.
        % For robustness, unreliable correspondences are downweighted by the  
        % similarity score map during optimization. 
        (c) The output refined pose. 
    }
    \label{fig:highlight}
    \vspace{-3ex}
\end{figure}

%% file: doc/related_work.tex
\input{fig/framework.tex}

\textbf{6-DoF object pose estimation.} 
% Recent deep learning methods 
6-DoF object pose estimation systems (usually going after an object detector~\cite{he2017mask,tian2019fcos,carion2020end,yi2020segvoxelnet,zhu2020ssn,mao2021pyramid,yang2021pdnet,yang2022vgvl}) aim to estimate the 3-DoF orientations and 3-DoF locations of rigid objects. 
The boom of deep learning has significantly improved object pose estimation in recent years. 
Some methods proposed to directly regress object poses from monocular color images \cite{kehl2017ssd,XiangSNF18,li2019cdpn,hu2020single,chen2020learning,wang2021gdr} or with the aid from depth sensors \cite{wang2019densefusion,he2020pvn3d,wada2020morefusion}. They leveraged CNNs' regression ability to directly map the observed images to object poses. More recently, correspondence-based methods \cite{rad2017bb8,tekin2018real,park2019pix2pose,peng2019pvnet,hodan2020epos,chen2020end} become more popular. 
They employed CNNs to estimate the corresponding 3D model point for each observed object pixel, and then solve for poses with PnP \cite{fischler1981random}. 
These methods may estimate the object's bounding box corners \cite{rad2017bb8,tekin2018real}, predict dense 2D-3D correspondence maps\cite{park2019pix2pose} or vote the  keypoints by all object pixels \cite{peng2019pvnet}. 
More recently, EPOS \cite{hodan2020epos} proposed to 
handle symmetric objects by segmenting 3D models into patches and estimating the patch centers. 
The above direct object pose estimation methods usually become less stable when varying illuminations and occlusions exist.   
Several methods \cite{XiangSNF18,sundermeyer2018implicit,li2018deepim,manhardt2018deep,zakharov2019dpod,iwase2021repose} hence conducted pose refinement based on the estimated coarse initial pose above, which achieved significant performance gains. 
Some of these methods \cite{XiangSNF18,sundermeyer2018implicit} relied on depth data from costly sensors and utilized ICP 
% to refine the pose by 
to align the known object model to the observed depth image. While \cite{li2018deepim,manhardt2018deep,zakharov2019dpod} first rendered a 2D object image according to the initial pose  
% the object model as 
% 2D images according to the initial pose estimate 
and then compared the rendered image with the observed image via a CNN to estimate the residual pose. 
These RGB-based methods are especially attractive 
% the interest of the community 
due to their economical nature. 
However, most of these methods need massive training data and are not quite robust in practical scenarios. Moreover, they need a cumbersome CNN for pose regression, which sacrifices efficiency.  
% usually cannot attain real-time efficiency. 
% where the image feature extraction need to be done from scratch in each refinement iterations, which results in less efficient processing.  
Iwase \etal \cite{iwase2021repose} proposed to alleviate such dilemma by reusing the images features extracted by CNN and attained real-time processing. Concretely, they employed the CNN as an image feature encoder, based on which to formulate a non-linear optimization problem to align the features from the inference and target images for pose refinement inspired by BA-Net~\cite{tang2018ba}. 
Though efficient, their formulation is built upon % significant 
overlapped object regions across the reference image and target image, which may thus be less stable with erroneous initial pose inputs. 
The previous work \cite{grabner2020geometric} proposed to refine the pose based on the correspondences, but their method is still limited to ideal scenarios. % which is sensitive to the initial pose quality. 

\textbf{Non-linear least squares optimization with deep learning.}
Non-linear least squares optimization algorithms, such as Gauss-Newton~\cite{meloun2011statistical} and Levenberg-Marquardt~\cite{more1978levenberg}, are widely used in computer vision~\cite{mur2015orb,lin2021barf,xu2022robust,huang2021vs}, given their efficient and effective nature. 
Recently, the differentiability of the optimization algorithm itself has been widely studied and several works \cite{tang2018ba,von2020gn,teed2019deepv2d,sarlin2021back} have included the differentiable optimization algorithm during the network training for localization systems and visual SLAMs. These inspire our formulation for object pose refinement. 
% Our method shares spirits with these works but has made substantial modifications for the learning-based object pose estimation. 

% A leverage the CNN for pose refinement solely based on color images, which also achieve appealing performances without the need for other modality sensors. We mainly focus on the pose refinement with a RGB camera in this work, considering the economical RGB cameras are commonly available in most application scenarios.

%% file: fig/framework.tex
\begin{figure*}[t!]
    \centering
    \includegraphics[width=0.95\linewidth, trim=1cm 7.5cm 2cm 1.1cm, clip]{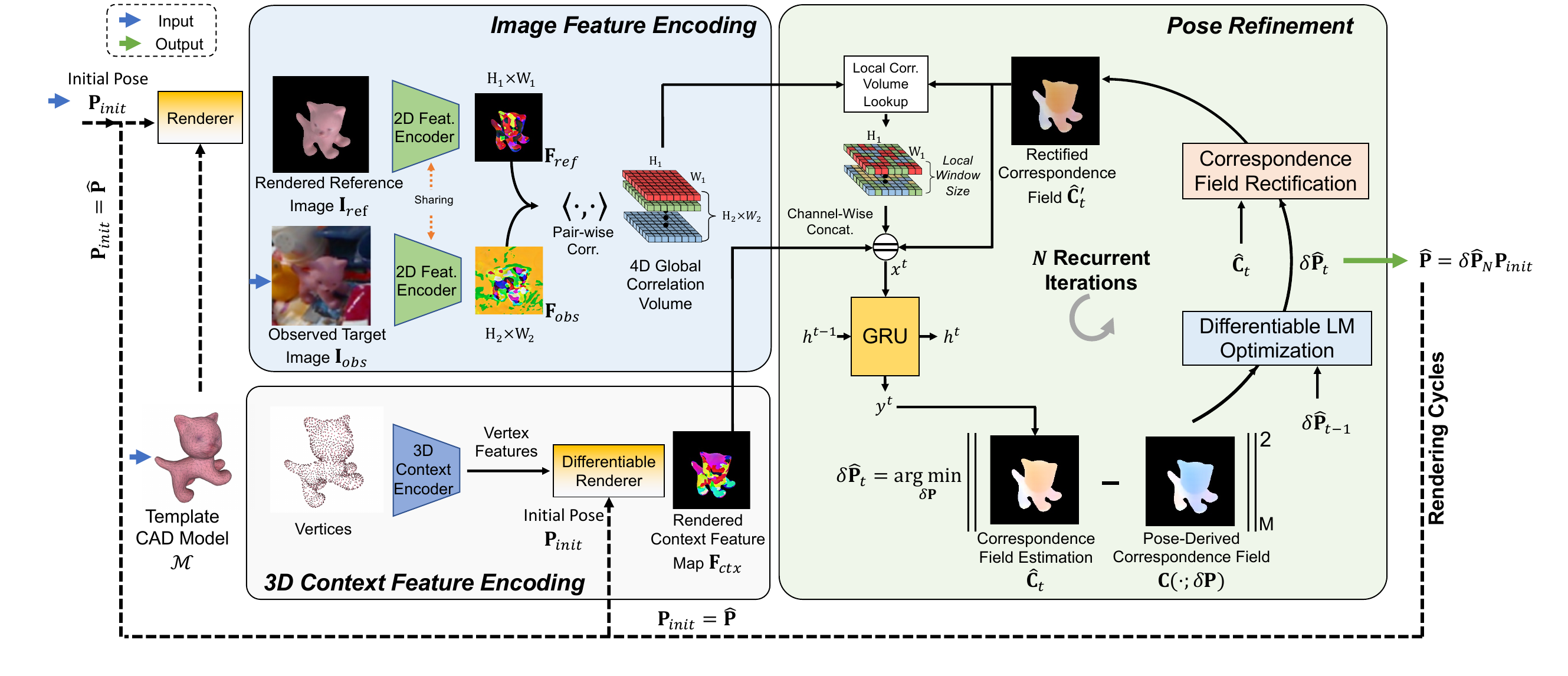}
    \vspace{-3ex}
    \caption{
        % Given an observed image $\mathbf{I}_{obs}$ and an initial object pose $\mathbf{P}_{init}$, RNNPose refines the initial pose to align with the observation recurrently. 
        Overview of the proposed method. 
        For pose refinement, a reference image $\mathbf{I}_{ref}$ is rendered with the object CAD model and its initial pose $\mathbf{P}_{init}$.   
        % Then, 
        The \textbf{image feature encoding} module encodes the rendered image $\mathbf{I}_{ref}$ and the observed image $\mathbf{I}_{obs}$ to feature maps and build a 4D global correlation volume. 
        % establish a pair-wise global correlation volume for lookup later.  
        In parallel, the \textbf{3D context feature encoding} module encodes the 3D model geometry and render the 3D features to a 2D context feature map $\mathbf{F}_{ctx}$ according to the initial pose estimation.
        %  to provide context information. % to align with $\mathbf{F}_{ref}$. 
        % which provides geometric context for the follow-up correspondence estimation. 
        % Based on the prerequisites above, 
        During \textbf{pose refinement}, 
        the correspondence field $\hat{\mathbf{C}}_t$
        and the residual pose $\delta\hat{\mathbf{P}}_t$ are alternately estimated in a recurrent framework. 
        % Based on the estimated field $\hat{\mathbf{C}}_t$ by a GRU, 
        % The Mahalanobis distance between $\mathbf{C}(\cdot; \delta\mathbf{P})$ (a correspondence field derived from the pose variable), and the estimated field $\hat{\mathbf{C}}_t$ by a GRU is minimized for pose estimation.   
        After the LM pose optimization, correspondence field estimation $\hat{\mathbf{C}}_t$ is rectified as $\hat{\mathbf{C}}_t'$ by enforcing rigid-transformation constraints with the currently optimized pose $\delta\hat{\mathbf{P}}_t$ to further improve next-iteration estimations.  
        % by enforcing rigid-transformation constraints. 
        After $N$ recurrent iterations, the reference image $\mathbf{I}_{ref}$ is re-rendered with the current pose estimation.
        %  for finer refinement.  
        %  to update the local correlation volumes for next iteration. 
    }
    \label{fig:framework}
    \vspace{-3ex}
\end{figure*}

%% file: doc/method.tex
% \vspace{-2ex}
Given an 
% observed target 
observed object image $\mathbf{I}_{obs}$, an initial object pose estimate $\mathbf{P}_{init}$ and the object's CAD model $\mathcal{M}$ as inputs, a 6-DoF pose refinement system aims to further improve the object pose estimation. 
% based on its initial pose $\mathbf{P}_{init}$.  
In this paper, we propose a recurrent pose refinement method, dubbed RNNPose, which is robust to erroneous initial poses and occlusions.  
Our method is based on a rendering pipeline and may have several rendering cycles as illustrated by Fig.~\ref{fig:framework}. 
% At the beginning of each rendering cycle, 
At the beginning of the first rendering cycle, a reference image  $\mathbf{I}_{ref}$ is rendered with the object's CAD model according to its initial pose $\mathbf{P}_{init}$ (estimated by any direct methods \cite{XiangSNF18,peng2019pvnet}). 
Then, the rendered reference image, the observed target image, and the vertices of the CAD model are encoded as high-dimensional features which will be used 
to estimate the correspondences 
% for correspondence estimation 
(between the rendered image and observed image) in the follow-up pose refinement module. 
The pose refinement module constitutes our major contribution, where we formulate 
% the refinement as 
an optimization problem based on the correspondence estimations. 
We integrate correspondence field estimation and pose refinement into a recurrent framework for robustness and efficiency. 
To handle occlusions, we generate point-wise distinctive descriptors for the 3D object model and observed images with a 3D-2D hybrid network, 
% (omitted in Fig.~\ref{fig:framework}), 
with which to downweight the unreliable correspondences during pose optimization. 
% we design a 3D-2D hybrid network (omitted in Fig.~\ref{fig:framework}) 
% % trained with a contrastive descriptor loss 
% to generate point-wise distinctive descriptors used to downweight the unreliable correspondence estimations during pose optimization. 
After every several recurrent iterations, the reference image $\mathbf{I}_{ref}$ is re-rendered with the currently optimized pose to decrease the pose gap to the target for the next cycle. 

In the ensuing subsections, we will detail the feature extraction (Sec. \ref{sec:feature_extraction}), recurrent pose refinement (Sec.~\ref{sec:reccurent_framework}), and the loss functions for training (Sec. \ref{sec:loss}).

\subsection{2D-3D Feature Encoding and Rendering}\label{sec:feature_extraction}
% \subsection{Feature Encoding and 3D Feature Rendering}\label{sec:feature_extraction}
The rendered reference image $\mathbf{I}_{ref}$ and observed target image $\mathbf{I}_{obs}$ first need to be encoded into high-dimensional feature maps $\mathbf{F}_{ref}$ and $\mathbf{F}_{obs}$ 
% to construct a correlation volume for correspondence field estimation
for the follow-up feature correlation volume construction~\cite{dosovitskiy2015flownet,ilg2017flownet,hui2018liteflownet,sun2018pwc,teed2020raft}. 
The correlation volume encodes the appearance similarities between image pixels, which is essential for correspondence reasoning.
In our work, we adopt several residual blocks~\cite{he2016deep} for image feature encoding, and the pair-wise correlations of the encoded features are calculated to create a global correlation volume. 
The global correlation volume will be frequently queried for correspondence field estimation in the follow-up pose refinement module.   

% for the follow-up correlation lookup during correspondence field estimation.
% a global correlation volume is established, which contains the feature correlations of the two inputs. The global correlation volume is created for the follow-up correlation lookup during correspondence field estimation.  

Besides the pair-wise correlation volume, popular dense correspondence estimation methods also incorporate context features of the reference image for guidance. % which is helpful. 
% especially for 
% ambiguous regions with repeated patterns 
% regions under different illuminations or those with no texture.
As shown in Fig.~\ref{fig:framework}, to better encode the geometric contexts, 
unlike previous methods encoding the context features from 2D images,  
we directly encode the features from 3D object point clouds with a \textit{3D context feature encoder} based on KPConv~\cite{thomas2019kpconv}. 
The point-wise geometric features are then rendered as a 2D context feature map $\mathbf{F}_{ctx}$ according to the initial object pose estimation.   
Here, we adopt a differentiable renderer \cite{ravi2020pytorch3d} for feature rendering to enable geometric feature learning. 
We empirically found that encoding the context features from point clouds brings more robustness. 
% leads to more robust performance. 
% can produce robust geometric features,  which 
% improves the system's robustness. 
Besides, the vertex features only need to be extracted once per object model and archived for inference after training, which is quite efficient.  

\subsection{Recurrent Correspondence Field Estimation and 6-DoF Pose Refinement}\label{sec:reccurent_framework}
Based on the constructed correlation volume and encoded context features, we propose a 6-DoF object pose refinement system by integrating the correspondence estimation and pose optimization as a recurrent framework. The correspondence field estimation and pose optimization rely on each other and improve recurrently for robust pose refinement. 
% for robustness.    
% inspired by RAFT~\cite{teed2020raft}, 
% to improve the robustness. % to large initial pose errors. 
The basic pipeline is illustrated in the pose refinement module in Fig.~\ref{fig:framework}. 

\vspace{-2ex}
\subsubsection{Correspondence Field Estimation}
% \vspace{-1ex}

For correspondence field estimation, we adopt a network architecture similar to RAFT \cite{teed2020raft} but make major modifications to suit our task,  
\ie, including the 3D context feature encoding (Sec.~\ref{sec:feature_extraction}) and correspondence rectification 
% with rigid-transformation constraints 
(Sec.~\ref{sec:CF_rectify}). 
At the beginning of each recurrent iteration, 
for each pixel of the reference image, we first look up 
and collect (from the global correlation volume) its correlation values with the candidate pixels in the target image. 
The candidate pixels are within a square local window centered at the estimated correspondences from the previous iteration. 
% for correspondence estimation. 
% The candidate pixels should reside in a square local window centered at the estimated correspondences from previous iteration. 
% (within a local window centered ) 
% we look up the global correlation volume to collect a local  correlation volume 
% (2D feature map)  
% for correspondence estimation. 
% The local correlation volume contains the correlations 
The collected correlations are then reshaped as a local correlation volume (a 2D map) spatially aligned with the reference image.
%  for correspondence estimation. 
%  and the reference image feature map are spatially aligned. 
% Each vector in the correlation volume contains the correlations of the reference image feature with a set of its correspondence candidates in the target image.  
% The correspondence candidates reside in a local window centered at the estimated corresponding point from the previous iteration. 
In the first iteration, 
% the correspondence field is initialized with zeros, 
% the correlation lookup is based on all-zero correspondence offsets, 
we use an all-zeros correspondence field to bootstrap correlation candidate identification,  while in the later iterations, the rectified correspondence field 
% with the currently refined pose 
(to be elaborated in Sec.~\ref{sec:CF_rectify}) is used.

After the correlation lookup, the collected local correlation volume, the rectified correspondence field, and the previously encoded context feature map $\mathbf{F}_{ctx}$ are concatenated as inputs to a GRU network to estimate the correspondence field $\hat{\mathbf{C}}_{t}$ for the current ($t$-th) recurrent iteration. 
% After the correlation lookup, the collected local correlation volume and the rectified correspondence field are concatenated with the previously encoded context feature map $\mathbf{F}_{ctx}$ as inputs to a GRU network. 
% The GRU estimates the correspondence field $\hat{\mathbf{C}}_{t}$  for the current ($t$-th) recurrent iteration based on these inputs, as shown in Fig.~\ref{fig:framework}. 

% \vspace{-2ex}
\subsubsection{6-DoF Pose Refinement} 
% \vspace{-1ex}
% \noindent\textbf{Pose Optimization Formulation.} 
\noindent\textbf{Basic Formulation.} 
\iffalse
Based on the rendered reference image (with pose $\mathbf{P}_{init}$) and the observed target image, the pose refinement module estimates the residual pose from the reference image to the target image
and rectify initial pose $\mathbf{P}_{init}$ with the estimation. 
\fi
Given a reference image (with depth map) 
and a target image, the 
ground-truth correspondence field of the reference image can be derived based on the ground-truth residual pose $\delta \mathbf{P}^{gt}$ point-wisely: 
% % with pose difference $\delta \mathbf{P}^{gt}$, 
% the correspondence of a reference object point $\mathbf{x}^i$ can be, 
% if the ground-truth pose difference $\delta \mathbf{P}^{gt}$ is known. 
% % between the reference image and target image is known. 
% By denoting a correspondence field as $\mathbf{C}$, the correspondence of a reference point $\mathbf{x}^i$ is derived as:   
%  and their pose difference (ground-truth) $\delta \mathbf{P}^{gt}$   
% $\mathbf{C}$ between the reference image and the target image 
% can be derived with their ground-truth pose difference $\delta \mathbf{P}^{gt}$. 
% The derived correspondence of each object point $\mathbf{x}^i\in \mathbb{R}^2$ in the reference image is expressed as
% between the rendered reference image and the target image can be derived with the ground-truth pose difference $\delta \mathbf{P}^{gt}$ as:
\begin{equation}\label{eq:pose_derived_field}
   \mathbf{C}(\mathbf{x}^i; \delta\mathbf{P}^{gt} ) = \pi(\delta\mathbf{P}^{gt} \pi^{-1}(\mathbf{x}^i, z^i)),
\end{equation}
where
$\mathbf{C}(\mathbf{x}^i; \delta\mathbf{P}^{gt} )\in\mathbb{R}^2$ denotes the ground-truth correspondence field value of point $\mathbf{x}^i$, and $z^i$ denotes the associated rendered depth value. Here, $\pi(\cdot)$ and $\pi^{-1}(\cdot; z^i)$ are the projection (3D-to-2D) and inverse projection (2D-to-3D) functions of a pinhole camera model.

To estimate the residual pose,  
% $\delta \hat{\mathbf{P}}\approx \delta\mathbf{P}^{gt}$,  
we take the correspondence field $\hat{\mathbf{C}}_t$ estimated by the GRU as an approximation of its ground-truth, \ie, $ \hat{\mathbf{C}}_t(\mathbf{x}^i) \approx\mathbf{C}(\mathbf{x}^i; \delta\mathbf{P}^{gt} )$, 
and 
push the correspondence field derived by the pose argument $\delta \mathbf{P}$, \ie, $ \mathbf{C}(\mathbf{x}^i; \delta\mathbf{P} )$, close to the GRU's estimation
% :  $ \mathbf{C}(\mathbf{x}^i; \delta\mathbf{P} ) \rightarrow \hat{\mathbf{C}}_t(\mathbf{x}^i) \approx\mathbf{C}(\mathbf{x}^i; \delta\mathbf{P}^{gt} )$ 
by optimizing $\delta \mathbf{P}$.
In this way, the residual pose parameter $\delta \mathbf{P}$ will approximate the ground-truth $\delta \mathbf{P}^{gt}$ after the optimization.   
The specific formulation is a non-linear least squares problem and the objective function is expressed as 
\begin{equation}\label{eq: energy_simple}
E(\bm{\xi} ) = \sum_{i=1}^M (\hat{\mathbf{C}}_t(\mathbf{x}^i) -\mathbf{C}(\mathbf{x}^i; \bm{\xi}) )^T (\hat{\mathbf{C}_t}(\mathbf{x}^i) -\mathbf{C}(\mathbf{x}^i; \bm{\xi})), 
\end{equation}
where the residual pose argument 
% $\delta\mathbf{P}\in SE(3)$ 
$\delta\mathbf{P}$ is parameterized as its minimal representation $\bm{\xi}\in\mathfrak{se}(3)$ (of the associated Lie-algebra) during optimization. 
% $\hat{\mathbf{C}}_{t}(\mathbf{x}^i)$ 
$\hat{\mathbf{C}}_{t}$ is the GRU-estimated correspondence field at the $t$-th recurrent iteration, and $\mathbf{C}(\mathbf{x}^i; \bm{\xi})$ denotes the correspondence of point $\mathbf{x}^i$ derived with the pose parameter argument $\bm{\xi}$, and $M$ is the total number of object points in the rendered reference image.

\input{fig/corr_window}
\noindent\textbf{Handling Unreliable Correspondences with Similarity Scores.}\label{sec:similarity_score}
The formulation of Eq.~\eqref{eq: energy_simple} is based on an impractical assumption that the correspondence field $\hat{\mathbf{C}}_t$ can be reliably estimated for all foreground regions, which is extremely difficult considering ubiquitous occlusions. 
% which is impractical because of the ubiquitous occlusions. 
% The formulation of Eq.~\eqref{eq: energy_simple} assumes the correspondence field estimation $\hat{\mathbf{C}}_t$ is reliable for all foreground points which is impractical because of the ubiquitous occlusions. 
% varying illuminations \etc. 
We further propose to incorporate a consistency-check mechanism to downweight the unreliable values in $\hat{\mathbf{C}}_t$ during pose optimization. 
To model the reliability of estimated correspondence,  one option is to adopt a forward-and-backward consistency check \cite{liu2019selflow,meister2018unflow}. 
% following previous self-supervised methods~\cite{liu2019selflow,meister2018unflow}. 
However, the bidirectional consistency check doubles the computational cost, and the domain gap between the rendered images and the real images increases the learning difficulty. 

We therefore propose a descriptor-based consistency check to alleviate the dilemma. The basic idea is to represent the 3D object model $\mathcal{M}$ 
and the observed 2D target image $\mathbf{I}_{obs}$ as two sets of  distinctive descriptors point-wisely via a 3D-2D hybrid network (with KPConvs \cite{thomas2019kpconv} and a keypoint description net~\cite{detone2018superpoint} as backbones). 
% \footnote{
%    % Our hybrid network is 
%    Designed with KPConv \cite{thomas2019kpconv} and an efficient keypoint detection backbone\cite{detone2018superpoint}. 
% }.
The corresponding descriptors of the object model and object images % across these two sets 
are enforced to be similar, while the non-corresponding descriptors are enforced to be dissimilar (by training with a contrastive descriptor loss function being described in Sec.~\ref{sec:loss}). 
The learned 3D model descriptors are rendered as 2D feature maps, denoted as  $\mathbf{D}_{\mathcal{M}}$, according to the object pose of the reference image  
% (as in the context feature encoding module) 
for fast indexing. The encoded target image descriptor map is denoted as $\mathbf{D}_\mathcal{I}$.  

With these high-dimensional distinctive descriptors, for each estimated correspondence pair ($\mathbf{x}^i$, $\hat{\mathbf{C}}_t(\mathbf{x}^i)$), we measure its reliability according to the similarity between their associated 3D and 2D descriptors ($\mathbf{d}_\mathcal{M}^{i}$, $\mathbf{d}_\mathcal{I}^{i}$ ). $\mathbf{d}_\mathcal{M}^{i}$ and $\mathbf{d}_\mathcal{I}^{i}$ here are collected from the above descriptor maps: $\mathbf{d}_\mathcal{M}^{i}=\mathbf{D}_{\mathcal{M}}(\mathbf{x}^i)$ and $\mathbf{d}_\mathcal{I}^{i}=\mathbf{D}_\mathcal{I}(\hat{\mathbf{C}}_t(\mathbf{x}^i))$, where bilinear interpolation may be applied for non-inetger correspondence coordinates. 
% Specifically, we first collect the associated 3D model descriptor $\mathbf{d}_\mathcal{M}^{i}$ of source point $\mathbf{x}^i$, as well as the target image descriptor $\mathbf{d}_\mathcal{I}^{i}$ of the estimated corresponding point $\hat{\mathbf{C}}_t(\mathbf{x}^i)$. 
\iffalse
Specifically, we first collect the associated 3D model descriptor $\mathbf{d}_\mathcal{M}^{i}$ of source point $\mathbf{x}^i$, as well as the target image descriptor $\mathbf{d}_\mathcal{I}^{i}$ of the estimated corresponding point $\hat{\mathbf{C}}_t(\mathbf{x}^i)$. 
\fi
% Then, 
The reliability of this correspondence pair is modeled with a similarity score:
\begin{equation}\label{eq: similarity_score}
   w^i= \exp\left(- \frac{|1-\mathbf{d}_\mathcal{M}^{i^T} \mathbf{d}_{\mathcal{I}}^i|}{\sigma} \right),  
\end{equation} 
where $\sigma$ is a learnable parameter (initialized with $1$) adjusting the sharpness. 
The similarity scores are used as the weights of the Mahalanobis distance measurements in Eq.~\eqref{eq: energy_simple}, which effectively downweight unreliable correspondences during optimization. 
By introducing a diagonal weighting matrix $\mathbf{w}^i=\big(\begin{smallmatrix}w^i & 0 \\0 & w^i \end{smallmatrix}\big)$, the weighted version of Eq.~\eqref{eq: energy_simple} is written as 
\begin{equation}\label{eq: energy_weighted}
   E(\bm{\xi} ) = \sum_{i=1}^M (\hat{\mathbf{C}}_t(\mathbf{x}^i) -\mathbf{C}(\mathbf{x}^i; \bm{\xi}) )^T \mathbf{w}^i (\hat{\mathbf{C}}_t(\mathbf{x}^i) -\mathbf{C}(\mathbf{x}^i; \bm{\xi})). 
\end{equation}
% With this objective function, 
The pose optimization is thus formulated as
\begin{equation}\label{eq: optimization_formulation}
   \hat{\bm{\xi}} = \mathop{\arg\min}\limits_{\bm{\xi}} E(\bm{\xi}),  
   \vspace{-1ex}
\end{equation}
where the pose parameter $\bm{\xi}\in\mathfrak{se}(3)$ is optimized by minimizing the objective function defined by Eq.~\eqref{eq: energy_weighted}. 
% minimizing the distance between the two correspondence fields. 

\noindent\textbf{Differentiable Residual Pose Optimization. }
We solve the non-linear least squares problem (Eq.~\eqref{eq: optimization_formulation}) with Levenberg-Marquardt (LM) algorithm. 
For the optimization in the $t$-th recurrent iteration, the pose parameter is initialized with the estimated pose from the previous iteration \ie, $\bm{\xi}_0 = \log(\delta\hat{\mathbf{P}}_{t-1})$. 
% Starting from the pose estimation from the previous ( recurrent iteration $t-1$, \ie, $\bm{\xi}_0 = \log(\delta\hat{\mathbf{P}}_{t-1})$, 
Continuing from the parameter $\bm{\xi}_{p-1}$ of the previous LM iteration, the left-multiplied increment $\bigtriangleup\bm{\xi}_p$ 
% for the next update 
is computed by   
\begin{equation}\label{eq:lm_update}
      \bigtriangleup\bm{\xi}_p = (\mathbf{J}^T \mathbf{W} \mathbf{J}+\lambda\mathbf{I} )^{-1}\mathbf{J}^T \mathbf{W} \mathbf{r}(\bm{\xi}_{p-1}), 
      % (\hat{\mathbf{C}}(\mathbf{x}^i) - \mathbf{C}(\mathbf{x}^i; \bm{\xi} ) )  
      % (\hat{\mathbf{c}}^{i}-\mathbf{c}^i(\bm{\xi} ) ), 
\end{equation}
with which we update the parameter as $\bm{\xi}_{p} \leftarrow \bigtriangleup\bm{\xi}_{p} \circ \bm{\xi}_{p-1}$, to approach the optimal solution. 
Here, $\mathbf{J} = -\frac{\partial \mathbf{r}}{\partial\bm{\xi}}$ is the Jacobian matrix containing the derivative of the stacked residual vector $\mathbf{r}=(r_1, r_2, ..., r_{2M})^T$ (established from Eq.~\eqref{eq: energy_weighted})
with regard to a left-multiplied increment. 
% where $n$ is the number of object points in the reference image and each point corresponds to a 2D correspondence field value difference.
We unroll the parameter update procedure and make the LM optimization layer differentiable to enable end-to-end network training. The differentiable optimization procedure enhances the feature learning for correspondence field estimation, which is essential to high performance. 
% After several updates, 
After LM optimization, the residual pose of the $t$-th recurrent iteration is estimated as 
$\delta\hat{\mathbf{P}}_t = \exp(\hat{\bm{\xi}} )$,  where $\hat{\bm{\xi}}$ denotes the optimized parameter after several updates with Eq.~\eqref{eq:lm_update}. 

\noindent\textbf{Correspondence Field Rectification.}\label{sec:CF_rectify}
The erroneous initial poses usually produce large offsets between the rendered reference object and the observed object, which poses challenges for correspondence estimation.  
Moreover, unlike the standard scenarios of optical flow estimation~\cite{ilg2017flownet,hui2018liteflownet,sun2018pwc,teed2020raft}, unpatterned objects and varying illuminations are ubiquitous in our task, which further increases the difficulty. 
% in robust dense correspondence estimation.
Considering the optimized pose by Eq.~\eqref{eq: optimization_formulation} is mainly supported by the reliable correspondence estimations with our weighting mechanism  Eq.~\eqref{eq: similarity_score}, we 
% propose rectifying 
rectify the 
% the flawed values in the 
correspondence field as  $\hat{\mathbf{C}}'_t(\mathbf{x}) = \pi(\delta \hat{\mathbf{P}}_t \pi^{-1}(\mathbf{x}; z))$ based on the currently optimized pose $\delta \hat{\mathbf{P}}_t$. 
The rectification enforces the rigid-transformation constraints among the correspondence field, 
% based on $\delta \hat{\mathbf{P}}_t$, 
which improves the overall correspondence quality for the correlation volume lookup in the following recurrent iteration. 
A toy example is shown in Fig.~\ref{fig:corr_win} for better understanding. 

\noindent\textbf{Object Pose Estimation Update.}
% As illustrated in Fig.~\ref{fig:framework}, 
After every $N$ recurrent iterations, the residual pose is estimated as $\delta\hat{\mathbf{P}}_N$ by the RNN. We update the object pose estimation with the estimated residual pose $\delta\hat{\mathbf{P}}_N$ as  $\hat{\mathbf{P}}\leftarrow\delta\hat{\mathbf{P}}_N \mathbf{P}_{init}$,   
and we re-render the reference image $\mathbf{I}_{ref}$ based on this updated pose to start the next N-recurrent-iteration refinement, as illustrated in Fig.~\ref{fig:framework}. 
We refer to the N-recurrent-iteration refinement as a \textit{rendering cycle}, and the initial pose $\mathbf{P}_{init}$ for the next cycle is set to $\hat{\mathbf{P}}$ accordingly. 
The performance and efficiency with different rendering cycles and recurrent iterations will be discussed in Sec.~\ref{sec:ablation_rec_ren_iter}. 

% \subsection{Loss Functions and Training Details}\label{sec:loss}
\subsection{Loss Functions}\label{sec:loss}
\vspace{-1ex}
\noindent\textbf{Model Alignment Loss.} 
To supervise the residual pose estimations $\{\delta \hat{\mathbf{P}}_t | t=1\ldots N \}$ generated in each rendering cycle (including $N$ recurrent iterations), we apply these residual poses as the left-multiplied increments to the initial pose $\mathbf{P}_{init}$, having the corresponding object pose estimations $\{ \hat{\mathbf{P}}_t | t=1\ldots N \}$, where $\hat{\mathbf{P}}_t = \delta \hat{\mathbf{P}}_t \mathbf{P}_{init}$.   
Thereafter, we adopt a 3D model alignment loss to supervise these pose estimations for each rendering cycle:  
\begin{equation}
    L_{ma}=\sum_{t=1}^N||\hat{\mathbf{P}}_t\mathbf{X}_{model} - \mathbf{P}^{gt}\mathbf{X}_{model} ||_1, 
    % L_{ma}=\sum_{k=1}^K \sum_{t=1}^N||\hat{\mathbf{P}}_{k,t}\mathbf{X}_{model} - \mathbf{P}^{gt}\mathbf{X}_{model} ||_1, 
\vspace{-1ex}
\end{equation}
% where $\hat{\mathbf{P}}_{k,t}$ denotes the obtained object pose after rectifying the 
where $\hat{\mathbf{P}}_t$ is the object pose estimation mentioned above and $\mathbf{P}^{gt}$ denotes the ground-truth pose. Here, $\mathbf{X}_{model}\in \mathbb{R}^{4\times M}$ contains homogeneous coordinates of the $M$  model points. This loss function encourages the pose estimation to be close to the ground-truth so that the transformed model points can be well aligned.

\noindent\textbf{Correspondence Loss.} 
% For correspondence field estimation, 
We adopt $L1$ loss \cite{teed2020raft} for correspondence field supervision, where the ground-truth correspondence fields are derived 
% from the object model and ground-truth pose 
with Eq.~\eqref{eq:pose_derived_field} based on ground-truth poses. 

\input{fig/pose_vis}
\noindent\textbf{Descriptor Loss.}
We use circle loss $\mathcal{L}_{cir}$~\cite{sun2020circle} as the contrastive loss to supervise the point-wise descriptor learning of the 3D object model and the target images for similarity score calculation Eq.~\eqref{eq: similarity_score}. 
Concretely, we view the target image $\mathbf{I}_{obs}$ as two parts, \ie, the foreground region (object region) denoted as $fg(\mathbf{I}_{obs})$
and the background region denoted as $bg(\mathbf{I}_{obs})$. For each foreground descriptor 
$\bm{d}_\mathcal{I}^i\in fg(\mathbf{I}_{obs})$, 
we first find a set of its corresponding 3D descriptors $\{\bm{d}_\mathcal{M}^j\}_+$ of object model via KNN searching (see supplementary materials for details).  
\iffalse
$\mathcal{M}$ via KNN searching, where the foreground pixels are lifted with the ground-truth depth and transformed to the 3D object canonical frame with the ground-truth pose. 
\fi
Then, $\bm{d}_\mathcal{I}^i\in fg(\mathbf{I}_{obs})$ is enforced to be similar to  
% its 
% correspondences 
$\{\bm{d}_\mathcal{M}^j\}_+$ and dissimilar to the remaining non-corresponding descriptors  $\{\bm{d}_\mathcal{M}^k\}_- $ with circle loss $\mathcal{L}_{cir}$~\cite{sun2020circle}, 
% by the circle loss 
which is expressed as ${\mathcal{L}_{cir}}( \bm{d}_\mathcal{I}^i, \{\bm{d}_\mathcal{M}^j\}_+, \{\bm{d}_\mathcal{M}^k\}_- )$. 
Moreover, for background descriptors $\bm{d}_\mathcal{I}^i\in bg(\mathbf{I}_{obs})$ , we constrain them to be similar to each other in the background, while to be dissimilar to the foreground descriptor set $fg(\mathbf{I}_{obs})$ with loss  
% $\mathcal{L}_{cir}( \bm{d}_\mathcal{I}^i, \{\bm{d}_\mathcal{I}^j\}_+, \{\bm{d}_\mathcal{I}^k\}_- )$ 
$\mathcal{L}_{cir}( \bm{d}_\mathcal{I}^i, bg(\mathbf{I}_{obs}), fg(\mathbf{I}_{obs}) )$.  
Traversing all target image descriptors $\mathbf{d}_\mathcal{I}^i$, the descriptor loss is calculated as  
\begin{align}
% \begin{equation}
\small
%   L_d = & \sum_{\bm{d}_\mathcal{I}^i\in fg(\mathbf{D}_\mathcal{I}) } 
%   % \underset{ \substack{ \bm{d}_\mathcal{M}^j\in \mathcal{E}_p(\bm{d}_\mathcal{I}^i), 
%   % \\ \bm{d}_\mathcal{M}^k \in \mathcal{E}_n(\bm{d}_\mathcal{I}^i)}}
%     {\mathcal{L}_{cir}}( \bm{d}_\mathcal{I}^i, \{\bm{d}_\mathcal{M}^j\}_+, \{\bm{d}_\mathcal{M}^k\}_- )  \\ 
%     & + \sum_{\bm{d}_\mathcal{I}^i\in bg(\bm{D}_\mathcal{I}) } 
%   %  \underset{ \substack{\bm{d}_\mathcal{I}^j\in bg(\bm{D}_\mathcal{I})\setminus{\bm{d}_\mathcal{I}^i}, 
%   % \\\bm{d}_\mathcal{I}^k\in fg(\bm{D}_\mathcal{I})}}
%   {\mathcal{L}_{cir}}( \bm{d}_\mathcal{I}^i, \{\bm{d}_\mathcal{I}^j\}_+, \{\bm{d}_\mathcal{I}^k\}_- ). 
\begin{aligned}
   % \scriptsize
   L_d = & \sum_{\bm{d}_\mathcal{I}^i\in fg(\mathbf{I}_{obs}) } 
    {\mathcal{L}_{cir}}( \bm{d}_\mathcal{I}^i, \{\bm{d}_\mathcal{M}^j\}_+, \{\bm{d}_\mathcal{M}^k\}_- )  \\ 
    & + \sum_{\bm{d}_\mathcal{I}^i\in bg(\bm{I}_{obs}) } 
   {\mathcal{L}_{cir}}( \bm{d}_\mathcal{I}^i, bg(\mathbf{I}_{obs}), fg(\mathbf{I}_{obs}) )
 \end{aligned}
 \label{eq:descriptor_loss}
\end{align}
to supervise the descriptor learning. With the contrastive learning, the corresponding 2D-3D descriptors 
% 2D-3D corresponding point pairs 
% (of the target image and the 3D model)
would be similar while the noncorresponding ones would be dissimilar, 
% compared with the noncorresponding ones, 
% mismatched pairs, 
which provides the foundation for unreliable correspondence handling 
% in Eq.~\eqref{eq: energy_weighted} 
with similarity scores Eq.~\eqref{eq: similarity_score}.  

%% file: fig/corr_window.tex
\begin{figure}[t!]
    \centering
\includegraphics[width=0.85\linewidth, trim=1cm 18.5cm 23cm 2cm, clip]{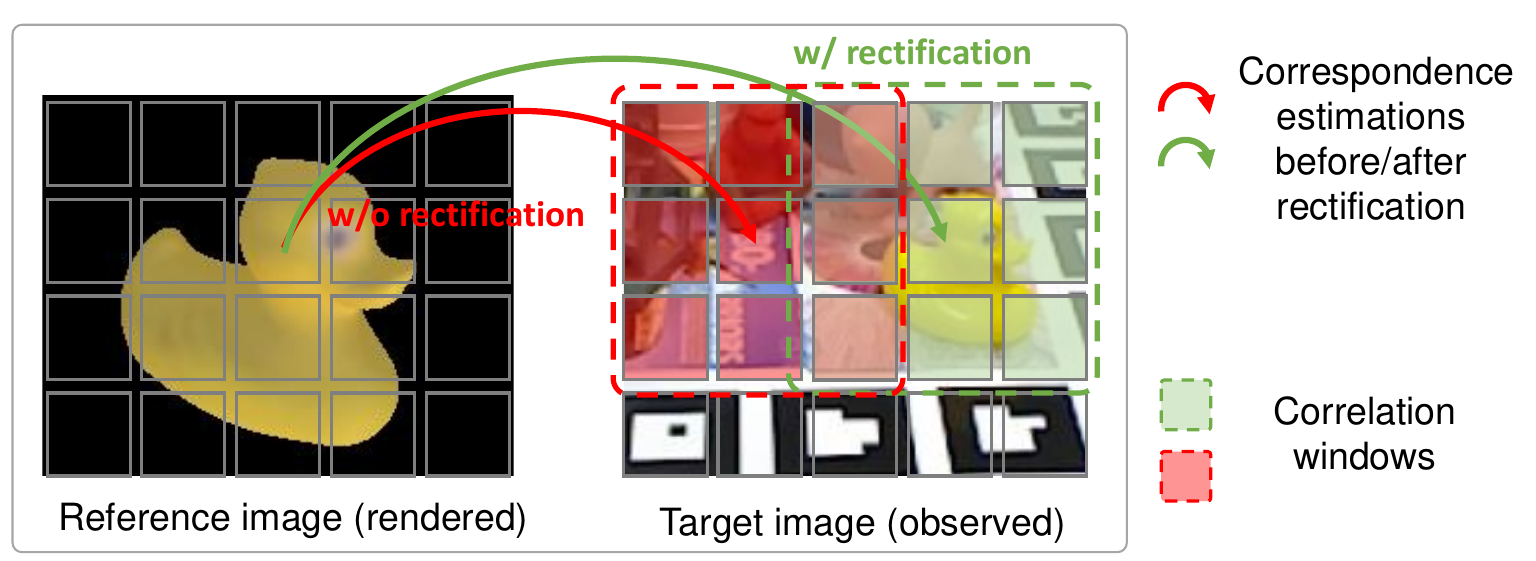}
    \vspace{-2ex}
    \caption{
        % A toy example to illustrate the differences of correlation window locations w/ or w/o correspondence field rectification. 
        With the rectified correspondences, 
        % the previous erroneous correspondences are rectified and 
        the related local correlation windows are accordingly shifted to better locations, which improves the estimation in the next recurrent iteration.  
        % A toy example to illustrate 
        % the differences of correlation window locations w/ or w/o correspondence field rectification. 
    % With rectification, the correlation window could better cover the target area for effective estimation. 
    }
    \label{fig:corr_win}
    \vspace{-3ex}
\end{figure}

%% file: fig/pose_vis.tex
\begin{figure*}[t!]
    \vspace{-1ex}
    \centering
    \includegraphics[width=0.99\linewidth, trim=0cm 17.5cm 2cm 0cm, clip]{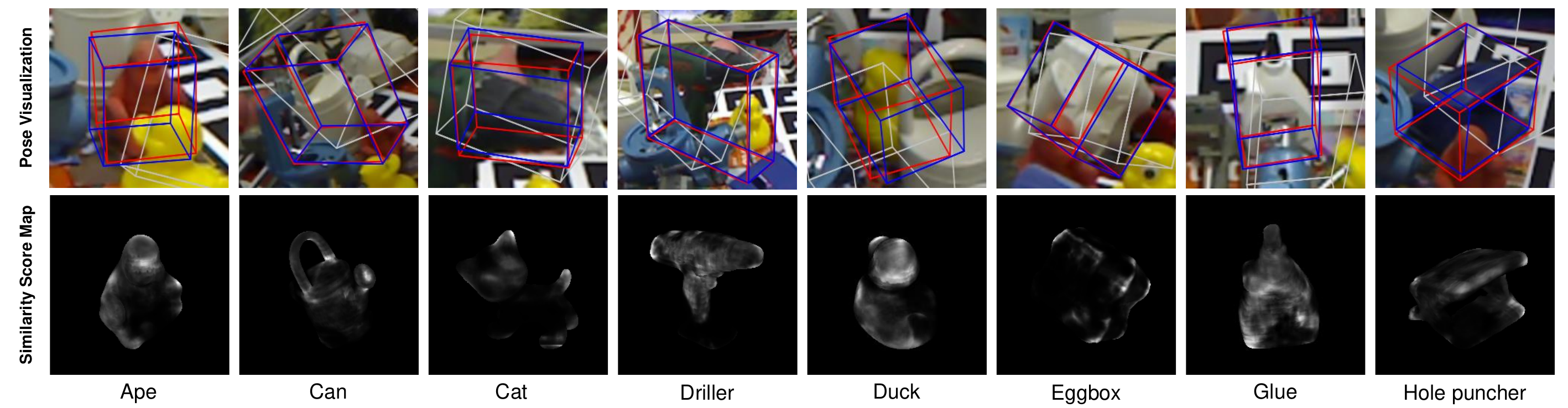}
    \vspace{-2ex}
    \caption{
        Visualization of our pose estimations (first row) on Occlusion LINEMOD dataset and the similarity score maps (second row) for downweighting unreliable correspondences during pose optimization. 
        For pose visualization, the white boxes represent the erroneous initial poses, the red boxes are estimated by our algorithm and the ground-truth boxes are in blue. Here, the initial poses for pose refinement are originally from PVNet~\cite{peng2019pvnet} but added with significant disturbances for robustness testing. % as in Fig.~\ref{fig:renreciter}. 
    }
    \label{fig:pose_vis}
    \vspace{-3ex}
\end{figure*}

%% file: doc/experiment.tex
\input{table/ablation_lm}

\subsection{Experimental Setup}
\noindent\textbf{Implementation Details.}
We train all of our networks end-to-end using the Adam~\cite{kingma2014adam} optimizer with an initial learning rate of $10^{-4}$ and adjust it with a cosine annealing strategy. The weights of model alignment loss $L_{ma}$ and descriptor loss $L_d$ are set to $1$, while the correspondence loss weight is set to $0.5$. During training, we conduct 3 rendering cycles, each of which performs 4 recurrent refinement iterations for pose refinement.   
All our models are trained agnostic to the initial pose sources where disturbed ground-truth poses are taken as initial poses for training following~\cite{li2018deepim}.
For testing, we conduct the same numbers of rendering cycles and refinement iterations as those during training for most experiments if without further declaration, though more iterations could produce better results. Please refer to the supplementary materials for more details.

\noindent\textbf{Datasets.}
We evaluate our method on three datasets, including LINEMOD~\cite{hinterstoisser2012model}, Occlusion LINEMOD~\cite{brachmann2014learning} and YCB-Video~\cite{XiangSNF18}. LINEMOD is a standard benchmark for 6D object pose estimation. This dataset contains texture-less objects in cluttered scenes captured with challenging illuminance variations. The Occlusion LINEMOD is a subset of LINEMOD dataset with additional annotations for occluded objects, which is suitable for testing the robustness to severe occlusions. Besides, the YCB-Video dataset contains the images of the YCB object set \cite{calli2015ycb} where strong occlusions, clutters are exhibited. It includes more than 110k real images captured for 21 objects with or without textures. We follow similar conventions in data processing and synthetic data generation as the previous works~\cite{peng2019pvnet,iwase2021repose}. For the initial poses, we mainly rely on PoseCNN~\cite{XiangSNF18} and PVNet~\cite{peng2019pvnet}, two typical direct estimation methods, following \cite{li2018deepim} and \cite{iwase2021repose}. We also create a set of extremely erroneous initial poses by adding random Gaussian noise to the original initial pose estimations to evaluate the robustness to large initial poses errors. 

\noindent\textbf{Evaluation Metrics.}
We evaluate our method with the metrics ADD-(S)~\cite{hinterstoisser2012model} and AUC of ADD(-S)~\cite{XiangSNF18}. 
For the ADD-(S) metric, the mean distance between the model points transformed with the pose estimation and the ground-truth is calculated. 
With the standard ADD-(S) metric,  if the mean distance is less than $10\%$ of the model diameter, the pose estimation is regarded as correct. In some of our experiments, we also test the performances when the threshold is set to $2\%$ or $5\%$ of the model diameter for stricter testing. For symmetric objects, the mean distance is computed based on closest point distances~\cite{hinterstoisser2012model}. 
When evaluating on the YCB-Video dataset, we also compute the AUC (Area Under Curve) of ADD(-S) by varying the distance threshold 
from 0~cm to 10~cm 
% with step size 0.1 cm 
following
% ~\cite{peng2019pvnet}. 
~\cite{XiangSNF18}. 

\subsection{Ablation Study}\label{sec:ablation}

\input{fig/robust_rriter_runtime.tex}
We conduct a thorough ablation study on LINEMOD and Occlusion LINEMOD datasets to evaluate the effectiveness of the components in our framework. 

\noindent\textbf{Correspondence Field Supervision.} 
We first remove the correspondence loss to verify the influence of correspondence field quality on the pose estimation.  
% We find the pose estimation is unstable and the model training crashes unless the rigid-transformation constraints are enforced by our correspondence rectification. 
The results `w/o correspondence loss' in Table \ref{tab:ablation} correspond to this ablation study, and the performance degrades significantly. Since our pose optimization is directly based on the correspondence field estimation, solid supervision on correspondence field estimation is essential to the overall system. 

\noindent\textbf{Effectiveness of the Pose Supervision and End-to-end Learning.} 
We further remove the supervision to the pose estimation by setting the weight of the model alignment loss $L_{ma}$ to 0. 
This is equivalent to adopting a typical non-differentiable LM optimizer because no gradient is backpropagated through the LM layer during training. It can be found that the object pose can still be reasonably estimated (denoted as `w/o $L_{ma}$' in Table \ref{tab:ablation}), but with humble performance, especially with stricter evaluation criteria, \ie, by setting a smaller threshold 0.01d or 0.05d. 
The performance degradation reflects the importance of end-to-end pose learning. The differentiable LM layer enables the pose supervision to affect the feature learning for more robust correspondence field estimation, which is essential to our formulation. 
% pose optimization formulation. 

\noindent\textbf{Correspondence Field Rectification. } 
Another key procedure in our recurrent pose refinement is the correspondence field rectification. To validate the effectiveness, we ablate this step and directly use the correspondence estimation $\hat{\mathbf{C}}_t$ from the GRU as the initialization for the next iteration (denoted as `w/o $\hat{\mathbf{C}}_t'$ rect.' in Table~\ref{tab:ablation}). We find that the performance drops significantly compared with our full framework, especially on more strict metrics, \ie, 0.01d and 0.05d. 
% The currently optimized pose is mostly based on the reliable correspondences with our similarity score weighting in Eq.~\eqref{eq: energy_weighted}. 
This phenomenon demonstrates that the corrected correspondence field with the rigid-transformation constraints from the optimized pose can facilitate the refinement in the following iterations.  
% By rectifying the correspondence field with the currently optimized pose,  unreliable values can be largely corrected, which facilitates the refinement in 
% the correlation volume quality for the 
% next recurrent iteration.   
% As the currently optimized pose is mostly based on the reliable correspondences via our similarity-score weighing (in Eq. \eqref{eq: energy_weighted}), rectifying the overall correspondence field with the optimized pose can thus correct the unreliable correspondences, with which local correlation windows can be positioned closer to true correspondence locations in the next iteration for better estimation. 

\noindent\textbf{3D Context Encoder.} 
To verify the effectiveness of our 3D context encoder, we test the system without the context encoder (denoted as `w/o 3D context $\mathbf{F}_{ctx}$') or with a commonly used 2D context encoder (denoted as `w/ 2D context'). The performances of these two versions both degrade compared to that with a 3D context encoder.
%  the candidate with a 2D context encoder performs relatively better.  
% than the version without a context encoder. 
The degradation not only reveals the importance of context information as indicated by previous works~\cite{sun2018pwc, teed2020raft}, but also proves that our 3D context encoder is a more effective choice than the 2D counterpart in our task. We reckon that the more robust performance may be attributed to the finer granularity of dense 3D point cloud features (compared with the low-resolution 2D image features). 
The finer-granularity features could provide more detailed geometric contexts. 
% while 2D features are learned from low-resolution 2D images. 
% It also shows that our 3D context encoder is an effective substitute for the 2D version, which is more robust in our task and 
% quite efficient for inference as the 3D encoder only needs to be run once per model to archive the features. 
\input{fig/robustness.tex}

\noindent\textbf{Similarity Scores for Occlusion Handling.} In 
% the right part of 
Table~\ref{tab:ablation}(b), we evaluate the effectiveness of similar scores in occlusion handling on the Occlusion LINEMOD dataset. 
The version `w/ similarity score' performs better for severely occluded objects. 
By including similarity scores during pose optimization, flawed correspondence estimations in the occluded unreliable regions are effectively downweighted.
%  to improve the robustness. 
 Some similarity score map examples are exhibited in Fig.~\ref{fig:pose_vis} for better understanding.   
% We further exhibit some pose estimation results and the similarity score maps in Fig.~\ref{fig:pose_vis}. 

\noindent\textbf{Recurrent Iterations vs Rendering Cycles. }\label{sec:ablation_rec_ren_iter}
The number of refinement iterations affects the system performance, especially when erroneous initial pose estimations exist. 
% More iterations generally produce better results but also with a heavier computational burden. 
We analyze the performances with different recurrent iterations and rendering cycles in Fig.~\ref{fig:renreciter}. 
% We visualize the performances with different recurrent iterations and rendering iterations in Fig.~\ref{fig:renreciter}. 
From Fig.~\ref{fig:renreciter}(a), it can be found that, by solely increasing the recurrent iterations while rendering the reference object image only once, we have achieved a high accuracy of $96.05\%$ which is comparable to RePOSE~\cite{iwase2021repose}. If conducting refinement with more recurrent iterations and rendering cycles, steady improvements are reported, which reflect good convergence of our method. To further validate the robustness to erroneous initial poses, we add Gaussian noise to the initial poses. Specifically, we randomly disturb translation components and rotation Euler angles with Gaussian noise. For the rotation, we add angular noise with standard deviation (STD, denoted as $\sigma_r$) of $10^\circ$ in all three axes.  
For the translational disturbance, we apply noise with a STD of $15$ cm along the $z$ axis (the axis perpendicular to the image plane) and 
% $5\times$ smaller deviations in 
STDs of $3$ cm in  
$x$ and $y$ directions ($\frac{1}{5}\times$) considering current methods usually have larger variances on depth estimations. 
From Fig.~\ref{fig:renreciter}(b), we find that the necessity of recurrent refinement becomes more noticeable.  

Though more rendering cycles bring performance gains as well, the extra costs are significant, since most of the input features need re-encoding. 
% compared with increasing recurrent iterations 
% since the reference image rendering, context feature rendering image feature encoding and the 2D part of the 2D-3D hybrid net all need to be rerun. 
% which need more computational budgets 
% According to the runtime analysis (Table~\ref{tab:runtime}), conducting more recurrent iterations is a more economical choice for better performance as only the CF (correspondence field) estimation, pose optimization and CF rectification modules are intensively activated. 
Based on the runtime analysis (Table~\ref{tab:runtime}), increasing the recurrent iterations is more economical for better performance as only the CF (correspondence field) estimation, pose optimization and CF rectification modules are activated for a recurrent iteration.  
% More detailed runtime analysis 
% Given the 6-DoF object pose estimation usually require real-time responses, re-rendering the reference inputs brings extra costs from the renderer, image feature encoder, 

\subsection{Comparison with State-of-the-Art Methods}
\input{table/eval_lm}
\input{table/eval_lmocc}

We compare with  
% state-of-the-art 
the cutting-edge methods on LINEMOD, Occlusion LINEMOD, and YCB-Video. 
%  datasets. 

For the LINEMOD dataset, we compare with the recent pose refinement methods RePOSE~\cite{iwase2021repose}, DPOD~\cite{zakharov2019dpod} and DeepIM~\cite{li2018deepim} as well as some direct estimation baselines~\cite{XiangSNF18,peng2019pvnet,song2020hybridpose}. 
Table~\ref{tab:result_of_linemod} contains the comparison results and we achieve a state-of-the-art performance. Interestingly, we achieve slightly better average performance when using PoseCNN~\cite{XiangSNF18} as the initial pose generator 
% compared with using 
rather than the PVNet~\cite{peng2019pvnet}, although the pose accuracy of PVNet is much better as exhibited in Table~\ref{tab:result_of_linemod}. This phenomenon reveals the good tolerance of our system to erroneous initial poses. To test our robustness to even larger initial pose errors, we add random Gaussian pose noises to the initial rotation and translation components separately for accuracy evaluation similar to those in Sec.~\ref{sec:ablation_rec_ren_iter}. 
Fig.~\ref{fig:noise_acc} plots the accuracy variations \wrt the disturbance magnitudes. Our method exhibits strong robustness and works reasonably even with extremely noisy initial poses. 

We also conduct comparisons on Occlusion LINEMOD. As shown in Table~\ref{tab:lmocc}, we outperform the cutting-edge method~\cite{iwase2021repose} by a significant margin ($51.6\nearrow60.65$), which manifests the system robustness to occlusions. We visualize some of our pose estimates from severely occluded images in the first row of Fig.~\ref{fig:pose_vis}, where the initial poses from PVNet are  disturbed with Gaussian noise like before ($\sigma_t=15$ cm, $\sigma_r=10^\circ$) to pose more challenges. 
% are further randomly disturbed with the same Gaussian noise parameter for Fig.~\ref{fig:renreciter}.  
It is shown that our system is capable of handling large initial pose errors even in highly occluded scenarios.  

Our additional evaluation on the YCB-Video dataset uses PoseCNN as the pose initializer, following the settings of RePOSE~\cite{iwase2021repose}. We compare with the refinement methods based on monocular color images. Our system still performs well on this large-scale complex dataset. We consistently improve the initial poses provided by PoseCNN~\cite{XiangSNF18}, and outperform the cutting-edge pose refinement method RePOSE in both metrics,  
% and also achieve much better performance in ADD(-S) metric than \cite{di2021so}, a method trained with massive data generated by advanced rendering technique, 
% \ie, physically-based rendering, 
as shown in Table~\ref{tab:ycb}. 

\input{table/eval_ycb}
% \subsection{Runtime Analysis}
% % \input{table/runtime.tex}
% We analyse the runtime of key components of our solution on a Tesla V100 GPU as listed in Table~\ref{tab:runtime}. 
% As only the correspondence field (CF) estimation, pose optimization and the CF rectification need to be conducted intensively, 
% even with naive python implementation, our system can achieve real-time processing speed. Considering the runtime for the pose optimization module can be further decreased (to $\sim 1$ ms) if adopting similar optimized implementation as RePose, our recurrent framework is therefore quite efficient.  
% \vspace{-1ex}
\section{Conclusions and Limitations}
\vspace{-1ex}
We have presented a recurrent framework for 6-DOF object pose refinement. A non-linear least squares problem is formulated for pose optimization based on the estimated correspondence field between the rendered image and observed image.  
Descriptor-based consistency checking is included to downweight unreliable correspondences for occlusion handling. 
% The occlusions are handled with a consistency check based on the learned distinctive 2D and 3D descriptors. 
Our method performs robustly against erroneous pose initializations and severe occlusions, which achieves state-of-the-art performances on public datasets. 

One limitation of our method is that the trained model is object-specific similar to many other works~\cite{li2018deepim,zakharov2019dpod,iwase2021repose}. 
Although for a novel object, only the pose refinement module needs further finetuning, the limited generalization ability to unknown objects is still undeniable. More detailed discussions are in the supplementary material. 
% our method is still limited to known objects (having been seen during training), though 
% for a novel object, only the pose refinement module needs finetuning to achieve good performance. 
% However, our method is still only applicable to scenarios where the target objects are known and accurate CAD models are provided. 
In the future, we plan to extend our method to handle unknown objects 
% or to reduce the dependence on CAD models 
for better generality.

\begin{spacing}{0.}
{
\scriptsize
\noindent\textbf{Acknowledgement}. 
This work is supported in part by Centre for Perceptual and Interactive Intelligence Limited, in part by the General Research Fund through the Research Grants Council of Hong Kong under Grants (Nos. 14204021, 14207319, 14203118, 14208619), in part by Research Impact Fund Grant No. R5001-18, in part by CUHK Strategic Fund.
}
\end{spacing}

%% file: table/ablation_lm.tex
\begin{table*}[t]
  \centering
  \vspace{-1ex}
  \caption{(a) Ablation study on LINEMOD dataset. 
%   (\textbf{left}) 
  (b) Validation of effectiveness of similarity score 
%   (\textbf{right}) 
  on the Occlusion LINEMOD dataset with the ADD(-S) metric. For more detailed comparison, the evaluations with different thresholds of ADD(-S) metric are conducted, \ie, $2\%$, $5\%$ and $10\%$ of the model diameter denoted as 0.01d, 0.05d, 0.1d respectively. 
  }\label{tab:ablation}
  \vspace{-2ex}
  \begin{subtable}[h]{0.7\linewidth}
  % \raggedright
  % \captionof{table}{Ablation study on LINEMOD dataset with the ADD(-S) score. The results are based on the initial poses provide by PoseCNN~\cite{xiang2017posecnn}. }
   \caption{}
  \setlength\tabcolsep{3 pt}
  \resizebox{1\linewidth}{!}{
    \begin{tabular}{c||ccc|ccc|ccc|ccc|ccc|ccc}
    \hline
         &\multicolumn{3}{c}{w/o correspondence loss} & \multicolumn{3}{c}{w/o $L_{ma}$} & \multicolumn{3}{c}{w/o $\hat{\mathbf{C}}_t'$ rect.}  & \multicolumn{3}{c}{w/o 3D context $\mathbf{F}_{ctx}$} & \multicolumn{3}{c}{w/ 2D context}& \multicolumn{3}{c}{Full(Ours)}\\\hline
    % Init. Pose  & -       & -       & -             & self-designed   & PVNet         &PoseCNN&PVNet\\ \hline
    Object    &0.02d &0.05d &0.1d         &0.02d &0.05d &0.1d   &0.02d         &0.05d  &0.1d            &0.02d &0.05d &0.1d                  &0.02d        &0.05d&0.1d         &0.02d &0.05d &0.1d \\\hline
    Ape       &1.29  &17.60 &61.23        &8.65  &36.03 &70.35  &4.40          &35.76  &74.51           &14.86 &50.48        &80.10          &12.19        &52.22&82.33        &$\bm{18.76}$&$\bm{57.14}$& $\bm{88.19}$  \\
    Benchvise &31.60 &87.08 &99.32        &58.14 &94.30 &99.71  &$\bm{79.56}$  &98.72  &$\bm{100.0}$    &72.26 &$\bm{99.13}$ &$\bm{100.0}$   &75.26        &98.37&99.81        &75.17&98.25& $\bm{100.0}$ \\ 
    Camera    &19.37 &70.89 &94.90        &45.13 &82.31 &95.95  &$\bm{56.72}$  &90.09  &97.91          &53.63 &90.69        &$\bm{98.73}$   &56.90        &$\bm{91.68}$&97.78  &55.39&89.12& 98.04  \\
    Can       &8.95 &77.88  &96.83        &32.65 &86.71 &98.76  &47.13         &94.37  &99.31           &53.25 &95.28 &$\bm{99.80}$   &$\bm{53.21}$ &$\bm{95.62}$&99.72 &$\bm{54.53}$&94.69& 99.31  \\
    Cat       &4.59  &28.39 &71.64        &25.24 &62.60 &92.81  &31.74         &75.76  &97.98           &32.34 &74.55        &96.71          &$\bm{36.81}$ &$\bm{79.15}$&$\bm{98.55}$ &36.43&74.85& 96.41   \\
    Driller   &40.25 &84.04 &92.57        &49.88 &88.50 &98.22  &59.81  &$\bm{96.43}$ &$\bm{99.70}$     &60.46 &95.34        &$\bm{99.70}$   &60.69        &95.54&99.41       &$\bm{62.44}$&95.44& $\bm{99.70}$    \\
    Duck      &5.62  &22.44 &69.08        &16.66 &47.46 &79.69  &19.18         &55.68  &87.01           &16.71 &57.37        &85.92          &25.19        &$\bm{63.62}$ &88.01&$\bm{25.82}$&61.13& $\bm{89.30}$    \\      
    Eggbox    &43.45 &89.81 &$\bm{99.65}$ &46.40 &87.12 &98.12  &52.64         &83.45  &97.65           &50.05 &81.03        &95.59          &54.51        &86.38 &96.36       &$\bm{59.06}$&$\bm{93.80}$& 99.53    \\
    Glue      &44.08 &93.57 &69.83        &10.67 &52.84 &92.29  &51.83         &93.95  &$\bm{99.87}$    &55.12 &94.40        &99.52          &54.14        &$\bm{95.71}$&$\bm{99.87}$ &$\bm{60.14}$&95.56&99.71         \\       
    Holep.    &6.26  &15.89 &51.95        &31.55 &65.55 &95.04  &32.81         &70.22  &96.53           &24.26 &66.51        &93.91          &20.61        &56.03 &91.04       &$\bm{35.68}$&$\bm{75.26}$& $\bm{97.43}$               \\       
    Iron      &42.33 &96.09 &99.08        &52.14 &95.48 &99.69  &62.46         &97.32  &99.59           &63.74 &97.45        &$\bm{100.00}$  &63.07        &$\bm{98.24}$&$\bm{100.0}$ &$\bm{68.03}$&98.16& $\bm{100.0}$              \\      
    Lamp      &33.87 &88.57 &98.98        &30.18 &81.90 &99.17  &45.95         &94.75  &99.81           &46.35 &93.76        &$\bm{99.81}$   &60.71        &94.43 &99.00 &$\bm{61.32}$&$\bm{94.91}$& $\bm{99.81}$             \\             
    Phone     &2.33  &18.53 &55.59        &31.08 &72.81 &95.36  &36.55         &82.74  &98.13           &39.66 &82.06        &97.26          &$\bm{42.68}$ &83.85&$\bm{98.39}$  &42.30&$\bm{83.95}$& $\bm{98.39}$       \\ \hline      
    Average   &21.85 &60.83 &85.27        &33.72 &73.35 &93.47  &44.68         &82.25  &96.00           &44.82 &82.93        &95.93          &47.44        &83.91 &96.17        &$\bm{50.39}$&$\bm{85.56}$& $\bm{97.37}$    \\ \hline
    \end{tabular}
  }
  \end{subtable}%
    \hfill
  \begin{subtable}[h]{0.28\textwidth}
    \centering
    \caption{}
  \setlength\tabcolsep{3 pt}
  \resizebox{1\linewidth}{!}{
    \begin{tabular}{c||ccc|cccccccc}
      \hline
            & \multicolumn{3}{c}{w/o similarity score}&  \multicolumn{3}{c}{w/ similarity score (Ours)}     \\
      \hline 
      Object  &0.02d &0.05d &0.1d   &0.02d &0.05d &0.1d  \\
      \hline 
      Ape         &$\bm{0.17}$    &8.97            &$\bm{38.63}$     &0.09          &$\bm{9.74}$    &37.18         \\
      Can         &7.29           &53.69           &85.50            &$\bm{7.79}$   &$\bm{56.01}$   &$\bm{88.07}$         \\ 
      Cat         &$\bm{1.60}$    &$\bm{11.71}$    &27.97            &$\bm{1.60 }$   &{11.63}   &$\bm{29.15}$         \\
      Driller     &13.76          &52.47           &78.42            &$\bm{14.58}$   &$\bm{59.80}$   &$\bm{88.14}$         \\
      Duck        &0.18           &$\bm{11.31}$    &47.77            &$\bm{0.26 }$   &11.13          &$\bm{49.17}$         \\ 
      Eggbox      &2.98           &25.96           &61.28            &$\bm{4.94 }$   &$\bm{38.47}$   &$\bm{66.98}$         \\ 
      Glue        &6.98           &35.22          &$\bm{65.01}$      &$\bm{10.52}$   &40.97          &63.79         \\
      Holep.      &0.08           &18.33           &59.83            &$\bm{0.42 }$   &21.42          &$\bm{62.76}$         \\\hline
      Average     &4.13           &27.21           &58.05            &$\bm{5.02 }$   &$\bm{31.15}$   &$\bm{60.65}$         \\     
      \hline
    \end{tabular}
  }
\end{subtable}
\vspace{-3ex}
\end{table*}

%% file: fig/robust_rriter_runtime.tex
\begin{figure}[!ht]
    \centering
    \begin{minipage}{0.68\linewidth}
        \centering
        % \resizebox{1\linewidth}{!}{
        \includegraphics[width=1\linewidth, trim=0cm 17.5cm 27cm 0cm, clip]{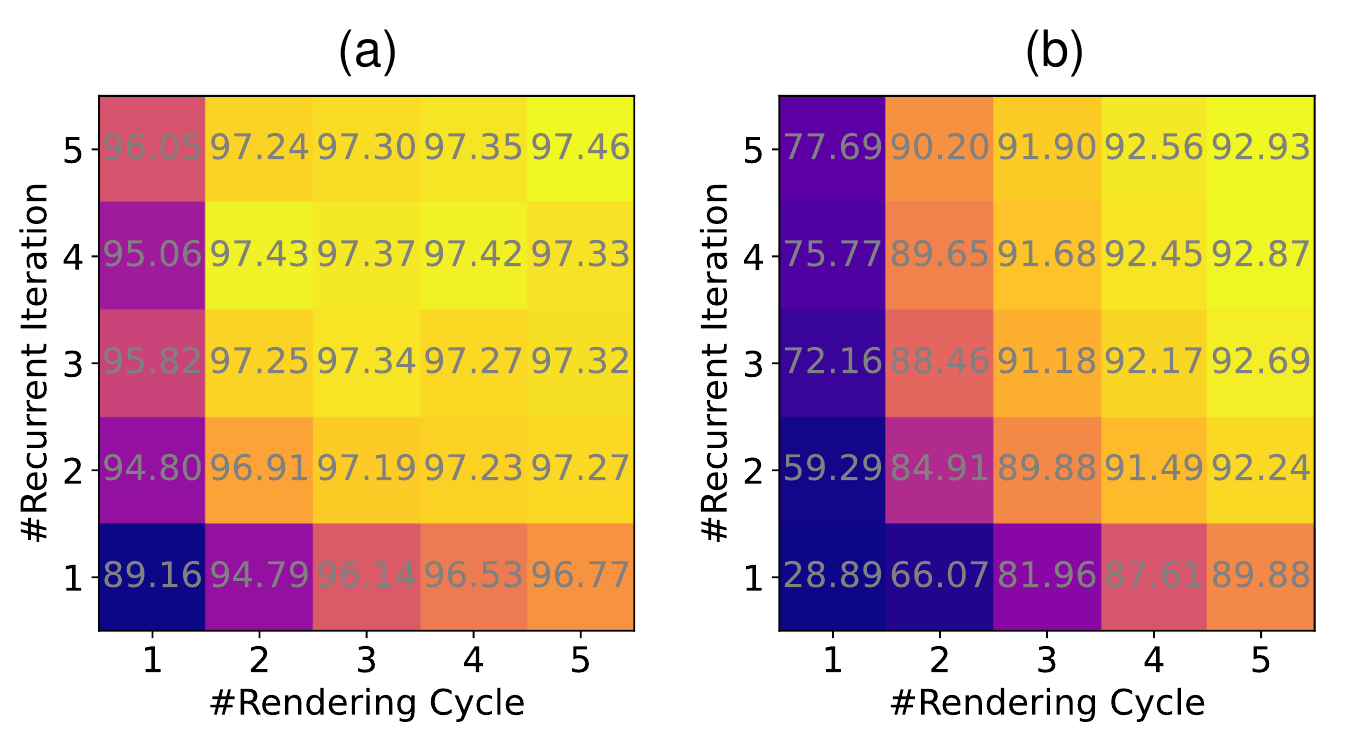}
        % }
        \vspace{-3ex}
        % \captionof*{figure}{ (I)
        % \label{fig:renreciter}
    \end{minipage}
    % \qquad
    \hfill
    \begin{minipage}{0.3\linewidth}
        \captionsetup{type=table} %% tell latex to change to table
        \vspace{0ex}
        % \caption{
        % % \captionof*{table}{ 
        %   Runtime analysis of individual modules.
        % }\label{tab:runtime}
        \setlength\tabcolsep{1pt}
        % \large
        % \Huge
        \centering
        \resizebox{1\linewidth}{!} {
        \begin{tabular}{c||cccccc}
        \hline
        Module & \begin{tabular}{@{}c@{}}Runtime \\(ms)\end{tabular}  \\
        \hline
        Ref. Image Rendering                                                 &8.88\\\hline
        \begin{tabular}{@{}c@{}}3D Context Encoding\\(run once per sequence) \end{tabular}     &35.20\\\hline
        \begin{tabular}{@{}c@{}}3D Feat. Rendering\\(context\&descriptor) \end{tabular}     &5.39\\\hline
        Image Feat. Encoding   &6.39\\\hline 
        \begin{tabular}{@{}c@{}}2D-3D Hybrid Net \\(2D part)\end{tabular} &2.99\\\hline
        CF Estimation                                                   &6.21\\ \hline
        Pose Optim.                                                     &6.23\\\hline
        CF Rectification                                                &1.48\\
        \hline
        \end{tabular}
        }
        % \label{tab:runtime}
        % \vspace{-4ex}
    \end{minipage}
    \captionlistentry[table]{A table beside a figure}\label{tab:runtime}
    % \captionlistentry[figure]{renreciter}\label{renreciter}
    \captionsetup{labelformat=andtable}
    \vspace{-1ex}
    \caption{
        \textbf{Left}: ADD(-S) accuracies \wrt different 
    % refinement iterations 
        recurrent iterations and rendering cycles on LINEMOD. 
        (a) Results based on the initial poses from PoseCNN~\cite{xiang2017posecnn} (b) Results based on the disturbed PoseCNN poses (with Gaussian noise $\sigma_t=15$cm, $\sigma_{r}=10^{\circ}$). \textbf{Right}: runtime analysis of individual modules.
        % with Gaussian noise ($\sigma_t=15$cm, $\sigma_{r}=10^{\circ}$). 
        % (a) Results with the initial poses from PoseCNN~\cite{xiang2017posecnn}. (b) Results with disturbed PoseCNN poses in (a) with Gaussian noise ($\sigma_t=15$cm, $\sigma_{r}=10^{\circ}$). 
    }\label{fig:renreciter}
    \vspace{-3ex}
\end{figure}

%% file: fig/robustness.tex
\begin{figure}[t!]
    \centering
    \vspace{-1ex}
    \includegraphics[width=0.9\linewidth, trim=0cm 17cm 18cm 0cm, clip]{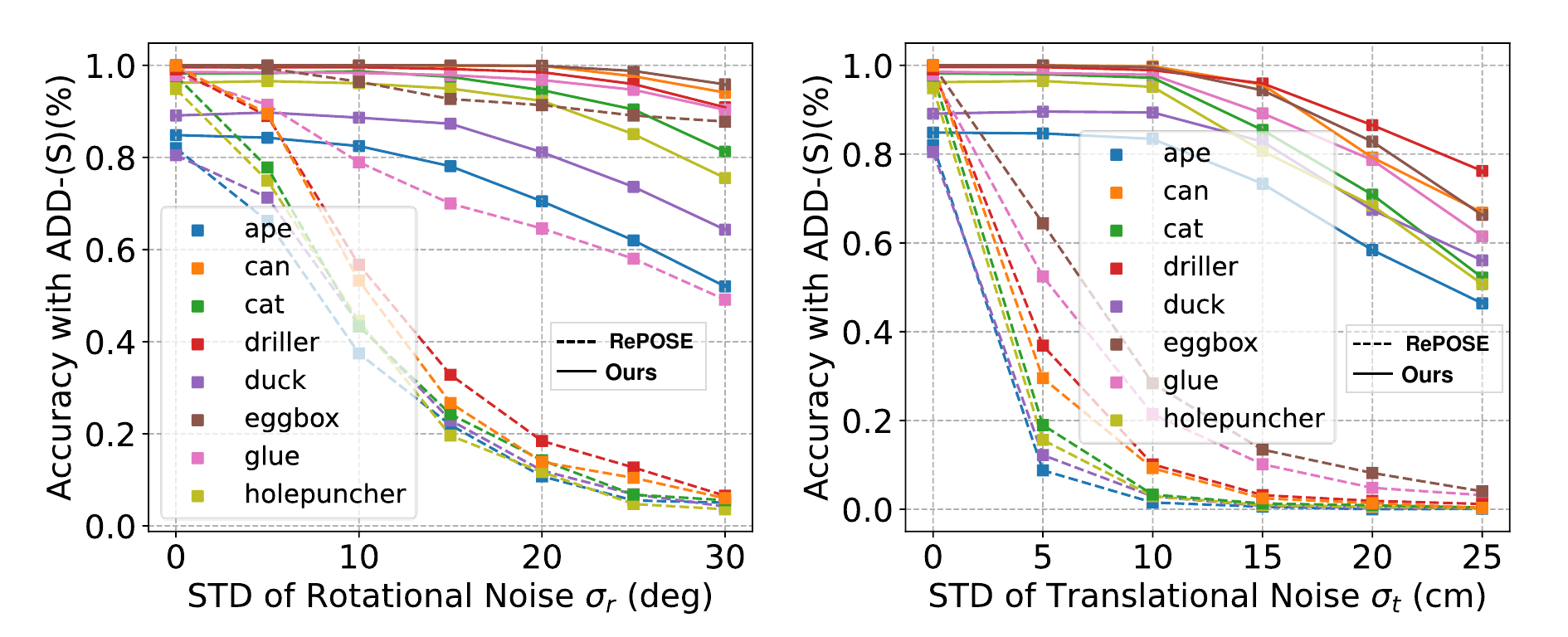}
    \vspace{-2ex}
    \caption{Robustness comparison with RePOSE by degrading the initial poses (from PVNet~\cite{peng2019pvnet}) with Gaussian noise on LINEMOD dataset.   }
    \label{fig:noise_acc}
    \vspace{-3ex}
\end{figure}

%% file: table/eval_lm.tex
\begin{table}[t]
  % \small
  \centering
  \caption{The comparison of estimation accuracy with competitive direct methods (PoseCNN~\cite{xiang2017posecnn}, PVNet~\cite{peng2019pvnet} and  HybridPose~\cite{song2020hybridpose}) and refinement methods (DPOD~\cite{zakharov2019dpod}, DeepIM~\cite{li2018deepim} and RePOSE\cite{iwase2021repose}) on LINEMOD dataset in terms of the ADD(-S) metric.
  }\label{tab:eval_lm}
  \vspace{-1.5ex}
  % \scalebox{0.8}{
  \resizebox{1\linewidth}{!}{
  \setlength\tabcolsep{1.5pt}
  \begin{tabular}{c||ccc|cccc|cccc}
    \hline
    Method      & PoseCNN & PVNet   & HybridPose  & DeepIM~\cite{li2018deepim}      & DPOD~\cite{zakharov2019dpod}           &\multicolumn{2}{c|}{RePOSE}~\cite{iwase2021repose} & \multicolumn{2}{|c}{Ours} \\ \hline
    Init. pose  &-        &-        &-             & PoseCNN      &self-designed     & PoseCNN &PVNet         & PoseCNN& PVNet            \\ \hline
    Ape         & 25.62   & 43.62    & 63.1         & 76.95        & 87.73             &47.4& 79.5            &$\bm{88.19}$  &85.62                           \\
    Benchvise   & 77.11   & 99.90    & 99.9         & 97.48        & 98.45             &88.5& $\bm{100.0}$   &$\bm{100.0}$  &$\bm{100.0}$                            \\
    Camera      & 47.25   & 86.86    & 90.4         & 93.53        & 96.07             &67.0& $\bm{99.2}$     &98.04  &98.43                           \\
    Can         & 69.98   & 95.47    & 98.5         & 96.46        & 99.71             &88.0& $\bm{99.8}$     &99.31  &99.51                           \\
    Cat         & 56.09   & 79.34    & 89.4         & 82.14        & 94.71             &80.6& $\bm{97.9}$     &96.41  &96.41                           \\
    Driller     & 64.92   & 96.43    & 98.5         & 94.95        & 98.80             &78.5& 99.0            &$\bm{99.70}$  &99.50                          \\
    Duck        & 41.78   & 52.58    & 65.0         & 77.65        & 86.29             &66.1& 80.3            &89.30  &$\bm{89.67}$                    \\
    Eggbox      & 98.50   & 99.15    & $\bm{100.0}$ & 97.09        & 99.91             &98.6& $\bm{100.0}$    &99.53  &$\bm{100.0}$                    \\
    Glue        & 94.98   & 95.66    & 98.8         & 99.42        & 96.82              &95.6& 98.3           &$\bm{99.71}$  &97.30                          \\
    Holep.      & 52.24   & 81.92    & 89.7         & 52.81        & 86.87             &62.7&96.9             &$\bm{97.43}$  &97.15                         \\
    Iron        & 70.17   & 98.88    & $\bm{100.0}$ & 98.26        & $\bm{100.0}$      &80.3& $\bm{100.0}$    &$\bm{100.0}$ &$\bm{100.0}$              \\
    Lamp        & 70.73   & 99.33    & 99.5         & 97.50        & 96.84             &87.8&  99.8           &99.81  &$\bm{100.0}$                     \\
    Phone       & 53.07   & 92.41    & 94.9         & 87.72        & 94.69             &74.3& $\bm{98.9}$     &98.39  &98.68                  \\ \hline
    Average     & 63.26   & 86.27    & 91.3         & 88.61        & 95.15             &78.1& 96.1            &$\bm{97.37}$  &97.10                   \\ \hline
    % \# of wins  & 0    & 1             & 2            & 6             & \textbf{8} \\ 
    \hline
  \end{tabular}
  }
  \vspace{-2ex}
  \label{tab:result_of_linemod}
  \end{table}
  

%% file: table/eval_lmocc.tex
\begin{table}[t]
  \centering
  \caption{
    Accuracy comparison with the state of the art on OCCLUSION LINEMOD dataset in terms of the ADD(-S) metric.
    }
    \vspace{-2ex}
  % \scalebox{0.8}{
  \setlength\tabcolsep{2pt}
  \resizebox{1\linewidth}{!}{
    \begin{tabular}{c||cccc|cc|c}
      \hline
      Object      & PoseCNN~\cite{xiang2017posecnn} & PVNet~\cite{peng2019pvnet}   & HybridPose~\cite{song2020hybridpose}  &GDR-Net~\cite{wang2021gdr} & DPOD~\cite{zakharov2019dpod}            & RePOSE~\cite{iwase2021repose}        &\multicolumn{1}{c}{Ours} \\\hline
      % Init. Pose  & -       & -       & -              &    & self-designed   & PVNet         &PoseCNN&PVNet\\ \hline
      Ape         & 9.60    & 15.8    & 20.9           &$\bm{39.3}$   & -        & 31.1 &  $\bm{37.18}$   \\
      Can         & 45.2    & 63.3    & 75.3           &79.2          & -        & 80.0 &  $\bm{88.07}$   \\
      Cat         & 0.93    & 16.7    & 24.9           &23.5          & -        & 25.6 &  $\bm{29.15}$   \\
      Driller     & 41.4    & 65.7    & 70.2           &71.3          & -        & 73.1 &  $\bm{88.14}$   \\
      Duck        & 19.6    & 25.2    & 27.9           &44.4          & -        & 43.0 &  $\bm{49.17}$   \\
      Eggbox      & 22.0    & 50.2    &52.4            &58.2          & -        & 51.7 &  $\bm{66.98}$   \\
      Glue        & 38.5    & 49.6    & 53.8           &49.3          & -        & 54.3 &  $\bm{63.79}$   \\
      Holep.      & 22.1    & 39.7    & {54.2}         &58.7          & -        & 53.6 &  $\bm{62.76}$   \\ \hline
      Average     & 24.9    & 40.8    & 47.5           &53.0          & 47.3     & 51.6 &  $\bm{60.65}$   \\ \hline
    \end{tabular}
    }
  \label{tab:lmocc}
  \vspace{-2ex}
\end{table}

%% file: table/eval_ycb.tex
\begin{table}[t]
  \caption{
    % Comparison with the methods based on single color images on the YCB-Video dataset. 
    Comparison with the refinement methods based on single images on the YCB-Video dataset. The performance of our initial pose generator, \ie, PoseCNN, is also included.  
    % Some of the results follow \cite{iwase2021repose} for comparison. 
    }
  \vspace{-1ex}
  \centering
  \resizebox{0.9\linewidth}{!} {
  \begin{tabular}{c||c|ccc|c}
  \hline
  Metric  & PoseCNN~\cite{xiang2017posecnn} & \multicolumn{1}{c}{DeepIM~\cite{li2018deepim}} 
  & DPOD~\cite{zakharov2019dpod} & {RePOSE}~\cite{iwase2021repose} &  Ours \\  \hline %& \multicolumn{1}{c}{CosyPose~\cite{labbe2020}}
  % AUC, ADD(-S) & 61.3 & 75.5 (81.9)  &$\bm{84.4}$   &80.8  &83.1 \\
  % AUC, ADD-S   & 75.2 & 83.1 (88.1)  &91.6   &86.7  &88.0  \\
  AUC, ADD(-S) & 61.3 & 81.9  &76.3  &80.8  &83.1 \\
  ADD(-S)      & 21.3 & 53.6  &50.4  &60.3  &{66.4}   \\ \hline
  \end{tabular}
  }
  \label{tab:ycb}
  \vspace{-3ex}
  \end{table}